\documentclass{llncs}
\usepackage[numbers,sort&compress,sectionbib]{natbib}
\usepackage{amsmath,graphicx}
\usepackage{amsmath,graphicx}
\usepackage[nolist]{acronym}
\usepackage{tabularx}
\graphicspath{{Figure/}}
\usepackage{color}
\usepackage{multirow}
\usepackage{siunitx}
\usepackage{booktabs} 
 \usepackage[font={small}]{caption, subfig}
\usepackage{tikz}

\newcolumntype{Y}{>{\centering\arraybackslash}X}
\newcommand{\ie}{\emph{i.e.},}
\newcommand{\eg}{\emph{e.g.},}

\newcommand{\etal}{\emph{et al.}}
\newcommand{\Sec}{\S}
\newcommand{\Fig}{Fig.}

\AtBeginDocument{}
\begin{acronym}
\acro{CT}{computed tomography}
\acro{HNN}{holistically-nested network}
\acro{P-HNN}{progressive holistically-nested network}
\acro{ILD}{interstitial lung disease}
\acro{LTRC}{Lung Tissue Research Consortium}
\acro{CNN}{convolutional neural network}
\acro{FCN}{fully convolutional network}
\acro{PLS}{pathological lung segmentation}
\acro{DS}{Dice score}
\acro{CNN}{convolutional neural network}
\acro{VS}{volumetric similarity}
\acro{COPD}{chronic obstructive pulmonary disease}
\acro{CV}{cross-validation}
\acro{NIH}{National Institutes of Health}
\acro{UHG}{University Hospitals of Geneva}
\acro{ILD}{interstitial lung disease}
\end{acronym}

\title{CT-Realistic Lung Nodule Simulation from 3D Conditional Generative Adversarial Networks for Robust Lung Segmentation}
\author{Dakai Jin \and Ziyue Xu\thanks{Corresponding author: ziyue.xu@nih.gov. This work is supported by NIH Intramural Research. We also thank NVidia for the donation of a Tesla K40 GPU.} \and Youbao Tang \and Adam P. Harrison \and Daniel J. Mollura  }
\institute{National Institutes of Health, Bethesda, MD, USA }

\begin{document}
\setlength{\abovecaptionskip}{1ex}
\setlength{\belowcaptionskip}{1ex}
\setlength{\floatsep}{1ex}

\maketitle

\begin{abstract}
Data availability plays a critical role for the performance of deep learning systems. This challenge is especially acute within the medical image domain, particularly when pathologies are involved, due to two factors: 1) limited number of cases, and 2) large variations in location, scale, and appearance. In this work, we investigate whether augmenting a dataset with artificially generated lung nodules can improve the robustness of the progressive holistically nested network (P-HNN) model for pathological lung segmentation of CT scans. To achieve this goal, we develop a 3D generative adversarial network (GAN) that effectively learns lung nodule property distributions in 3D space. In order to embed the nodules within their background context, we condition the GAN based on a volume of interest whose central part containing the nodule has been erased. To further improve realism and blending with the background, we propose a novel multi-mask reconstruction loss. We train our method on over $1000$ nodules from the LIDC dataset. Qualitative results demonstrate the effectiveness of our method compared to the state-of-art. We then use our GAN to generate simulated training images where nodules lie on the lung border, which are cases where the published P-HNN model struggles. Qualitative and quantitative results demonstrate that armed with these simulated images, the P-HNN model learns to better segment lung regions under these challenging situations. As a result, our system provides a promising means to help overcome the data paucity that commonly afflicts medical imaging. 
\end{abstract}


\section{Introduction}
\label{sec:intro}
Deep learning has achieved significant recent successes. However, large amounts of training samples, which sufficiently cover the population diversity, are often necessary to produce high quality results. Unfortunately, data availability in the medical image domain, especially when pathologies are involved, is quite limited due to several reasons: significant image acquisition costs, protections on sensitive patient information, limited numbers of disease cases, difficulties in data labeling, and large variations in locations, scales, and appearances.  Although efforts have been made towards constructing large medical image datasets, options are limited beyond using simple automatic methods~\citep{Karwoski2008}, huge amounts of radiologist labor~\citep{armato2011lung}, or mining from radiologist reports~\citep{wang2017chestx}. Thus, it is still an open question on how to generate effective and sufficient medical data samples with limited or no expert-intervention.

One enticing alternative is to generate synthetic training data. However, historically synthetic data is less desirable due to shortcomings in realistically simulating true cases. Yet, the advent of generative adversarial networks (GANs)~\citep{goodfellow2014generative} has made game-changing strides in simulating real images and data. This ability has been further expanded with developments on fully convolutional~\citep{radford2015unsupervised} and conditional~\citep{mirza2014conditional} GANs. In particular, Isola \etal{} extend the conditional GAN (CGAN) concept to predict pixels from known pixels~\citep{isola2017image}. Within medical imaging, Nie \etal{} use a GAN to simulate CT slices from MRI data~\citep{nie2017medical}, whereas Wolterink \etal{} introduce a bi-directional CT/MRI generator~\citep{wolterink2017deep}. For lung nodules, Chuquicusma \etal{} train a simple GAN to generate simulated images from random noise vectors, but do not condition based on surrounding context~\citep{chuquicusma2017fool}.

\begin{figure}[t]
\centering
\includegraphics[scale=.13]{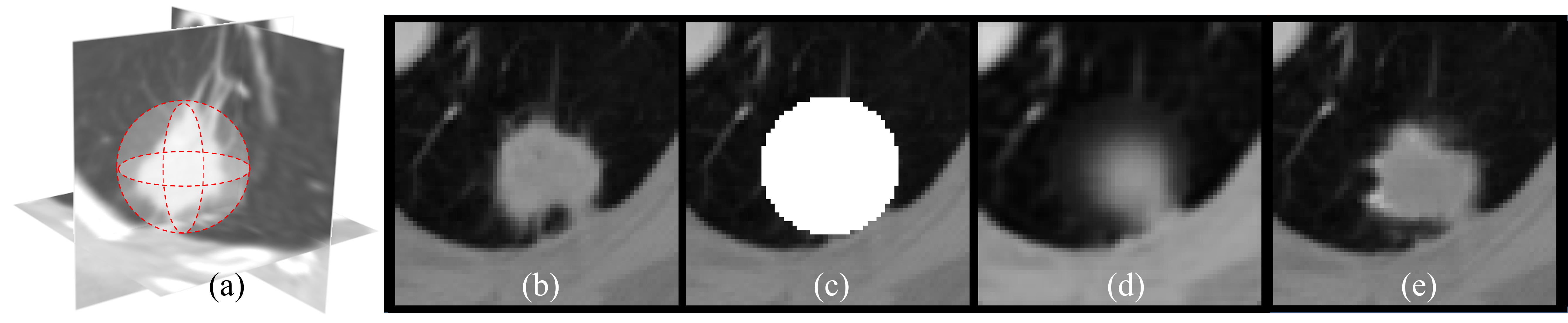}
\caption{Lung nodule simulation using the 3D CGAN. (a) A VOI centered at a lung nodule; (b) 2D axial view of (a); (c) same as (b), but with central sphere region erased; (d-e) simulated lung nodule using a plain $L1$ reconstruction loss and the 3D CGAN with multi-mask $L1$ loss coupled with adversarial loss, respectively. }
\label{fig:demo}
\end{figure}

In this work, we explore using CGAN to augment training data for specific tasks. For this work, we focus on pathological lung segmentation, where the recent  progressive holistically nested network (P-HNN) has demonstrated state of the art results~\citep{harrison2017progressive}. However, P-HNN can struggle when there are relatively large (\eg{} $>5mm$) peripheral nodules touching the lung boundary. This is mainly because these types of nodule are not common in Harrison \etal{}'s~\citep{harrison2017progressive} training set. To improve P-HNN's robustness, we generate synthetic 3D lung nodules of different sizes and appearances, at multiple locations, that naturally blend with surrounding tissues (see \Fig~\ref{fig:demo} for an illustration). We develop a 3D CGAN model that learns nodule shape and appearance distributions directly in 3D space. For the generator, we use a U-Net-like~\citep{Cicek2016} structure, where the input to our CGAN is a volume of interest (VOI) cropped from the original CT image with the central part, containing the nodule, erased (\Fig~\ref{fig:demo}(c)). We note that filling in this region with a realistic nodule faces challenges different than generating a random 2D nodule image from scratch~\citep{chuquicusma2017fool}. Our CGAN must generate realistic and natural 3D nodules conditioned upon and consistent with the surrounding tissue information. To produce high quality nodule images and ensure their natural blending with surrounding lung tissues, we propose a specific multi-mask reconstruction loss that complements the adversarial loss. 

The main contributions of this work are: (1) we formulate lung nodule generation using a 3D GAN conditioned on surrounding lung tissues; (2) we design a new multi-mask reconstruction loss to generate high quality realistic nodules alleviating boundary discontinuity artifacts; (3) we provide a feasible way to help overcome difficulties in obtaining data for "edge cases" in medical images; and (4) we demonstrate that GAN-synthetized data can improve training of a discriminative model, in this case for segmenting pathological lungs using P-HNN~\citep{harrison2017progressive}.

\section{Methods}
\label{sec:method}
\begin{figure}
\centering
\includegraphics[scale=.1435]{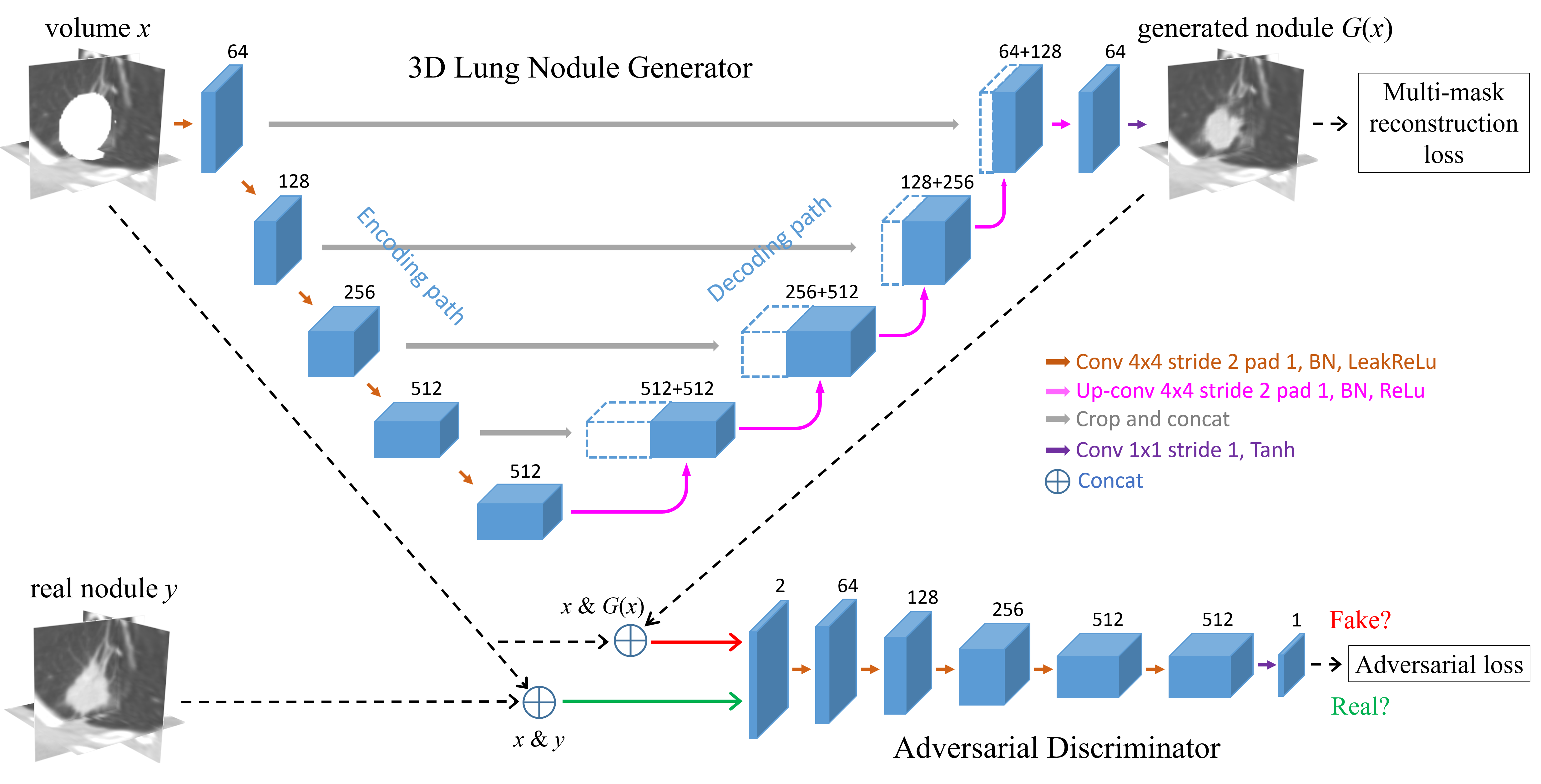}
\caption{3D CGAN architecture for lung nodule generation. The input is the original CT VOI, $y$, containing a real nodule and the same VOI, $x$, with the central region erased. Channel numbers are placed next to each feature map.}
\label{fig:gan_model}
\end{figure}

\Fig~\ref{fig:gan_model} depicts an overview of our method. Below, we outline the CGAN formulation, architecture, and training strategy used to generate realistic lung nodules.

\subsection{CGAN Formulation} 

In their original formulation, GANs~\citep{goodfellow2014generative} are generative models that learn a mapping from a random noise vector $z$ to an output image $y$. The generator, $G$, tries to produce outputs that fool a binary classifier discriminator $D$, which aims to distinguish real data from generated ``fake'' outputs. In our work, the goal is to generate synthetic 3D lung nodules of different sizes, with various appearances, at multiple locations, and have them naturally blend with surrounding lung tissues. For this purpose, we use a CGAN conditioned on the image $x$, which is a 3D CT VOI cropped from a specific lung location. Importantly, as shown in \Fig~\ref{fig:demo}(c), we erase the central region containing the nodule. The advantage of this conditional setting is that the generator not only learns the distribution of nodule properties from its surrounding context, but it also forces the generated nodules to naturally fuse with the background context. While it is possible to also condition on the random vector $z$, we found it hampered performance. Instead, like Isola \etal{}~\citep{isola2017image}, we use dropout to inject randomness into the generator.

The adversarial loss for CGANs can then be expressed as
\begin{align}
	\textit{\L}_{CGAN}(G,D)= \mathrm{E}_{x,y}[\log{D(x,y)}]+\mathrm{E}_{x}[\log{(1-D(x,G(x)))}]  \mathrm{,}
\end{align}
where $y$ is the original VOI and $G$ tries to minimize this objective against an adversarial discriminator, $D$, that tries to maximize it. Like others~\citep{pathak2016context, isola2017image}, we also observe that an additional reconstruction loss is beneficial, as it provides a means to learn the latent representation from surrounding context to recover the missing region. However, reconstruction losses tend to produce blurred results because it tends to average together multiple modes in the data distribution~\citep{isola2017image}. Therefore, we combine the reconstruction and adversarial loss together, making the former responsible for capturing the overall structure of the missing region while the latter learns to pick specific data modes based on the context. We use the $L1$ loss, since the $L2$ loss performed poorly in our experiments.

Since the generator is meant to learn the distribution of nodule appearances in the erased region, it is intuitive to apply $L1$ loss only to this region. However, completely ignoring surrounding regions during generator's training can produce discontinuities between generated nodules and the background. Thus, to increase coherence we use a new multi-mask $L1$ loss. Formally, let $M$ be the binary mask where the erased region is filled with $1$'s. Let $N$ be a dilated version of $M$. Then, we assign a higher $L1$ loss weight to voxels where $N-M$ is equal to one:
\begin{align}
	\textit{\L}_{L1}(G)= \mathrm{E}_{x,y}[M\odot\|y - G(x)\|_1] + \alpha \mathrm{E}_{x,y}[(N-M)\odot \|y - G(x)\|_1]  \mathrm{,} \label{eqn:both_l1}
\end{align}
where $\odot$ is the element-wise multiplication operation and $\alpha>=1$ is a weight factor. We find that a dilation of $3$ to $6$ voxels generally works well. By adding the specific multi-mask $L1$ loss, our final CGAN objective is
\begin{align}
	G^* = \arg\min_{G}\max_{D}\textit{\L}_{CGAN}(G,D) + \lambda \textit{\L}_{L1}(G)  \mathrm{,} \label{eqn:total_loss}
\end{align}
where $\alpha$ and $\lambda$ are determined experimentally. We empirically find $\alpha = 5$ and $\lambda = 100$ works well in our experiments.

\subsection{3D CGAN Architecture}

\Fig~\ref{fig:gan_model} depicts our architecture, which builds off of Isola \etal{}'s 2D work~\citep{isola2017image}, but extends it to 3D images. More specifically, the generator consists of an encoding path with $5$ convolutional layers and a decoding path with another $5$ de-convolutional layers where short-cut connections are added in a similar fashion to U-net~\citep{Cicek2016}. The encoding path takes an input VOI $x$ with missing regions and produces a latent feature representation, and the decoding path takes this feature representation and produces the erased nodule content. We find that without shortcut connections, our CGAN models do not converge, suggesting that they are important for  information flow across the network and for handling fine-scale 3D structures, confirmed by others~\citep{jin20173d}. To inject randomness, we apply dropout on the first two convolutional layers in the decoding path. 

The discriminator also contains an encoding path with $5$ convolutional layers. We also follow the design principles of Radford \etal{}~\citep{radford2015unsupervised} to increase training stability, which includes strided convolutions instead of pooling operations, LeakyReLu's in the encoding path of $G$ and $D$, and a Tanh activation for the last output layer of $G$.

\subsection{CGAN Optimization}
We train the CGAN model end-to-end. To optimize our networks, we use the standard GAN training approach \citep{goodfellow2014generative}, which alternates between optimizing $G$ and $D$, as we found this to be the most stable training regimen. As suggested by Goodfellow \etal{}~\citep{goodfellow2014generative}, we train $G$ to maximize $\log D(x, G(x))$ rather than minimize  $\log(1-D(x, G(x)))$.  Training employs the Adam optimizer~\citep{kingma2014adam} with a learning rate $0.0001$ and momentum parameters $\beta_1 = 0.5$ and $\beta_2 = 0.999$ for both the generator and discriminator.

\section{Experiments and Results}
We first validate our CGAN using the LIDC dataset~\citep{armato2011lung}. Then, using artificially generated nodules, we test if they can help fine-tune the state-of-the-art P-HNN pathological lung segmentation method~\cite{harrison2017progressive}.

\subsection{3D CGAN Performance}\label{sec:cgan_perf}

The LIDC dataset contains $1018$ chest CT scans of patients with observed lung nodules, totaling roughly $2000$ nodules. Out of these, we set aside $22$ patients and their $34$ accompanying nodules as a test set. For each nodule, there can be multiple radiologist readers, and we use the union of the masks for such cases. True nodule images, $y$, are generated by cropping cubic VOIs centered at each nodule with 3 random scales between $2$ and $2.5$ times larger than the maximum dimension of the nodule mask. All VOIs are then resampled to a fixed size of $64\times64\times64$. Conditional images, $x$, are derived by erasing the pixels within a sphere of diameter $32$ centered at the VOI. We exclude nodules whose diameter is less than $5mm$, since small nodules provide very limited contextual information after resampling and our goal is to generate relatively large nodules.  This results in roughly $~4300$ training sample pairs. We train the cGAN for $12$ epochs. 

\begin{figure}
\centering
\includegraphics[scale=.106]{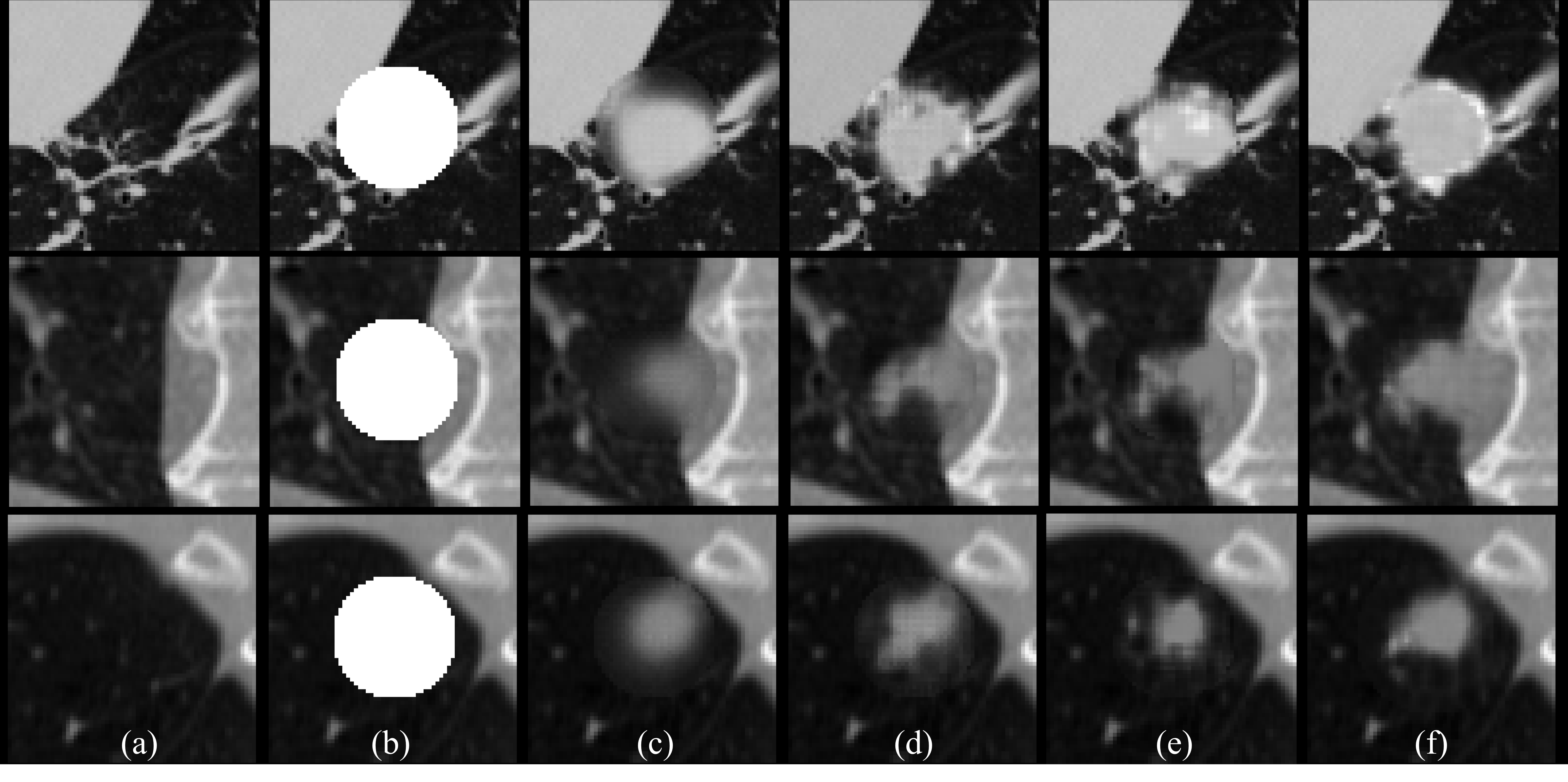}
\caption{Examples results: (a) original images; (b) input after central region erased; (c) only $L1$ loss, applied to the entire image; (d) Isola \etal{}'s method~\citep{isola2017image};  (e) CGAN with $L1$ loss, applied only to the erased region; (f) our CGAN with multi-mask $L1$ loss.}
\label{fig:gan_quality}
\end{figure}

We tested against three variants of our method: 1) only using an all-image $L1$ loss; 2) using both the adversarial and all-image $L1$ loss. This is identical to Isola \etal{}'s approach~\citep{isola2017image}, except extended to 3D; 3) using the same combined objective in \eqref{eqn:total_loss}, but not using the multi-mask version, \ie{} only using the first term of equation \eqref{eqn:both_l1}. As reconstruction quality hinges on subjective assessment~\citep{isola2017image}, we visually examine nodule generation on our test set. Selected examples are shown in \Fig~\ref{fig:gan_quality}.
As can be seen, our proposed CGAN produces realistic nodules of high quality with various shapes and appearances that naturally blend with surrounding tissues, such as vessels, soft tissue, and parenchyma. In contrast, when only using the reconstruction $L1$ loss, results are considerably blurred with very limited variations in shape and appearance. Results from Isola \etal{}'s method~\citep{isola2017image} improve upon the $L1$ only loss; however, it has obvious inconsistencies/misalignments with the surrounding tissues and undesired sampling artifacts that appear inside the nodules.  It is possible that by forcing the generator to reconstruct the entire image, it distracts the generator from learning the nodule appearance distribution. Finally, when only performing the $L1$ loss on the erased region, the artifacts seen in Isola \etal{}'s are not exhibited; however, there are stronger border artifacts between the $M$ region and the rest of the VOI. In contrast, by incorporating a multi-mask loss, our method can produce nodules with realistic interiors and without such border artifacts.

\subsection{Improving Pathological Lung Segmentation}

With the CGAN trained, we test whether our CGAN benefits pathological lung segmentation. In particular, the P-HNN model shared by Harrison \etal{}~\cite{harrison2017progressive} can struggle when peripheral nodules touch the lung boundary, as these were not well represented in their training set. Prior to any performed experiments, we selected $34$ images from the LIDC dataset exhibiting such peripheral nodules. Then, we randomly chose $42$ LIDC subjects from relatively healthy subjects with no large nodules. For each of these, we pick $30$ random VOI locations, centering within $(8,20)mm$ to the lung boundary with random size ranging $(32, 80)mm$. VOIs are resampled to 64$\times$64$\times$64 voxels and simulated lung nodules are generated in each VOI, using the same process as in \Sec\ref{sec:cgan_perf}, except the trained CGAN is only used for inference. The resulting VOIs are resampled back to their original resolution and pasted back to the original LIDC images, and then the axial slices containing the simulated nodules are used as training data ($\sim10000$ slices) to fine-tune the P-HNN model for $4-5$ epochs. For comparison, we also fine-tune P-HNN using images generated by the $L1$-only loss and also Isola \etal{}'s CGAN.
\begin{figure}
\centering
\includegraphics[scale=.38]{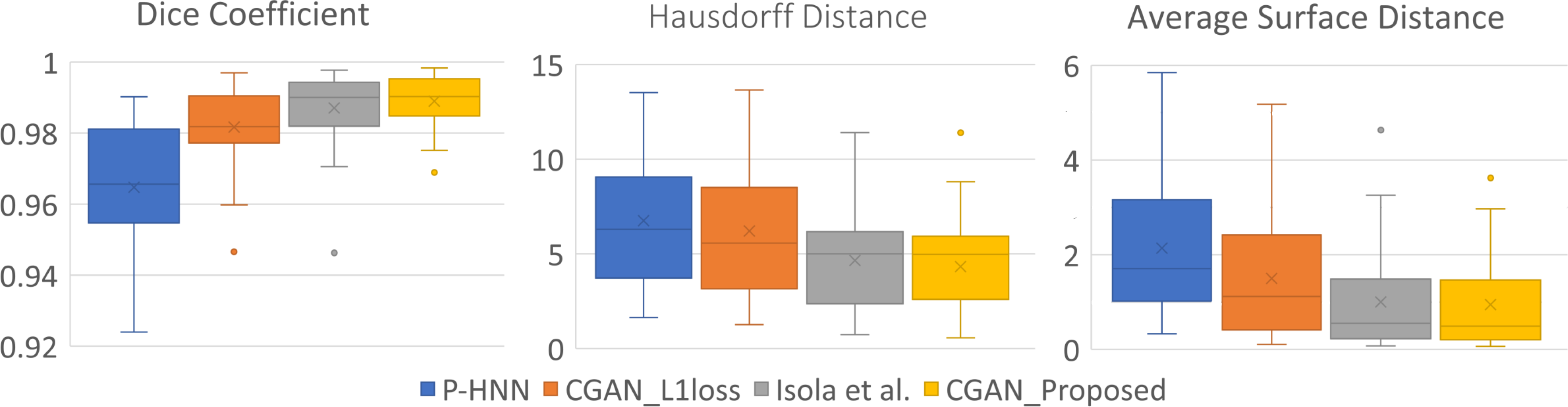}
\caption{Lung segmentation results on LIDC patients with peripheral lung nodules. All metrics are measured on a size $64$ pixel VOI centered on the nodule.}
\label{fig:seg_results}
\end{figure}
\begin{figure}
\centering
\includegraphics[scale=.192]{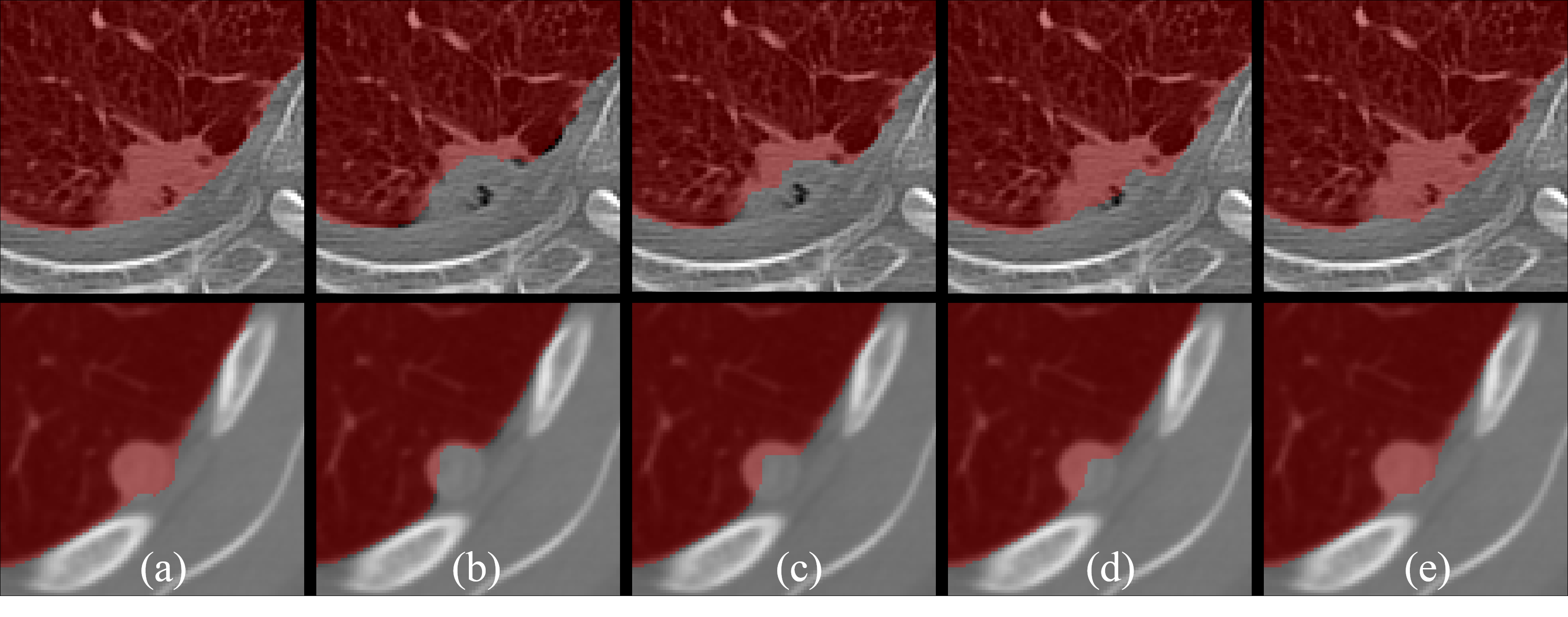}
\caption{Examples P-HNN lung segmentations. (a) ground truth; (b) original model; (c-e) fine-tuned with $L1$ loss only, Isola \etal{}~\citep{isola2017image}, and proposed CGAN, respectively.}
\label{fig:seg_quality}
\end{figure}

\Fig~\ref{fig:seg_results} depicts quantitative results. First, as the chart demonstrates, fine-tuning using \emph{all} CGAN variants improves P-HNN's performance on peripheral lung nodules. This confirms the value in using simulated data to augment training datasets. Moreover, the quality of nodules is also important, since the results using nodules generated by only an all-image $L1$ loss have the least improvement. Importantly, out of all alternatives, our proposed CGAN produces the greatest improvements in Dice scores, Hausdorff distances and average surface distances. For instance, our proposed CGAN allows P-HNN's mean Dice scores to improve from $0.964$ to $0.989$, and reduces the Hausdorff and average surface distance by $2.4mm$ and $1.2mm$, respectively. In particular, worse case performance is also much better for our proposed system, showing it can help P-HNN deal with edge cases. In terms of visual quality, \Fig~\ref{fig:seg_quality} depicts two examples. As these examples demonstrate, our proposed CGAN allows P-HNN to produce considerable improvements in segmentation mask quality at peripheral nodules, allowing it to overcome an important limitation. 

\section{Conclusion}
\label{sec:conclusion}

We use a 3D CGAN, coupled with a novel multi-mask loss, to effectively generate CT-realistic high-quality lung nodules conditioned on a VOI with an erased central region. Our new multi-mask $L1$ loss ensures a natural blending of the generated nodules with the surrounding lung tissues. Tests demonstrate the superiority of our approach  over three competitor CGANs on the LIDC dataset, including Isola \etal{}'s state-of-the-art method~\citep{isola2017image}. We further use our proposed CGAN to generate a fine-tuning dataset for the published P-HNN model~\cite{harrison2017progressive}, which can struggle when encountering lung nodules adjoining the lung boundary. Armed with our CGAN images, P-HNN is much better able to capture the true lung boundaries compared to both its original state and when it is fine-tuned using the other CGAN variants. As such, our CGAN approach can provide an effective and generic means to help overcome the dataset bottleneck commonly encountered within medical imaging.

\bibliographystyle{splncs}
\small{\bibliography{refs}}
\end{document}